\documentclass[10pt,twocolumn,letterpaper]{article}

\usepackage{cvpr}              % To produce the CAMERA-READY version
\usepackage[bottom]{footmisc}
\usepackage{colortbl}  % For cell colors
\usepackage{soul}
\usepackage[T1]{fontenc}
\usepackage{amsmath}
\usepackage{amsfonts}
\usepackage{multirow}
\usepackage{graphicx}
\usepackage{subcaption}
\usepackage[accsupp]{axessibility}
\usepackage{microtype}

\definecolor{cvprblue}{rgb}{0.21,0.49,0.74}
\usepackage[pagebackref,breaklinks,colorlinks,allcolors=cvprblue]{hyperref}

\def\paperID{0010} % *** Enter the Paper ID here
\def\confName{CVPR}
\def\confYear{2025}

\title{Physics-based Human Pose Estimation from a Single Moving RGB Camera}

\author{Ayce Idil Aytekin$^1$\thanks{Corresponding author: \texttt{aaytekin@mpi-inf.mpg.de}}\quad
Chuqiao Li$^2$\quad
Diogo Luvizon$^1$\quad
Rishabh Dabral$^1$\quad
Martin Oswald$^3$\\
Marc Habermann$^1$\quad
Christian Theobalt$^1$\\
$^1$ Max Planck Institute for Informatics\\
$^2$ University of Tübingen\\
$^3$ University of Amsterdam\\
}

\begin{document}

%% variables
\newcommand{\torque}{\boldsymbol{\tau}}
\newcommand{\jposition}{\mathbf{r}}
\newcommand{\jangle}{\mathbf{\theta}}
\newcommand{\shape}{\mathbf{\beta}}
\newcommand{\camera}{\mathbf{\pi}}
\newcommand{\camrot}{\mathbf{R_{GT}}}
\newcommand{\camtrans}{\mathbf{T_{GT}}}
\newcommand{\reel}{\mathbb{R}}
\newcommand{\pose}{\mathbf{q}}
\newcommand{\jacobian}{\mathbf{J}}
\newcommand{\grf}{\boldsymbol{\lambda}}
\newcommand{\inertia}{\mathbf{M}}
\newcommand{\nonlinear}{\mathbf{h}}
\newcommand{\roottrans}{\mathbf{q}_{:3}}
\newcommand{\deltat}{$\Delta t$}

\maketitle
\begin{abstract}
Most monocular and physics-based human pose tracking methods, while achieving state-of-the-art results, suffer from artifacts when the scene does not have a strictly flat ground plane or when the camera is moving.
Moreover, these methods are often evaluated on in-the-wild real world videos without ground-truth data or on synthetic datasets, which fail to model the real world light transport, camera motion, and pose-induced appearance and geometry changes. 
To tackle these two problems, we introduce MoviCam, the first non-synthetic dataset containing ground-truth camera trajectories of a dynamically moving monocular RGB camera, scene geometry, and 3D human motion with human-scene contact labels. 
Additionally, we propose PhysDynPose, a physics-based method that incorporates scene geometry and physical constraints for more accurate human motion tracking in case of camera motion and non-flat scenes. 
More precisely, we use a state-of-the-art kinematics estimator to obtain the human pose and a robust SLAM method to capture the dynamic camera trajectory, enabling the recovery of the human pose in the world frame. 
We then refine the kinematic pose estimate using our scene-aware physics optimizer.
From our new benchmark, we found that even state-of-the-art methods struggle with this inherently challenging setting, i.e. a moving camera and non-planar environments, while our method robustly estimates both human and camera poses in world coordinates. 
The code and the dataset will be released in \href{https://github.com/aidilayce/physdynpose}{https://github.com/aidilayce/physdynpose}.
\end{abstract}    
\section{Introduction}
\label{sec:intro}

Estimating accurate 3D human motion in global coordinates from a single moving RGB camera is an important and challenging problem in Computer Vision with many applications in animation, Augmented and Virtual Reality (AR/VR), human-robot interaction, autonomous driving, and assisted living environments.
This would, for example, allow to coherently track complex environment interactions within a metric-scale virtual scene.
%
%Therefore, to be able to estimate
Estimating precise human motion in global coordinates is essential for delivering a realistic and functional experience. 
%
%%%%%
%
\par 
However, most works so far have predicted 3D keypoints~\cite{Mehta2016Monocular3H, Mehta2017SingleShotM3}, joint angles~\cite{Kolotouros2019LearningTR, Li2020HybrIKAH}, or joint torques~\cite{Shimada2020PhysCap, Yuan2021SimPoESC} in a camera-relative coordinate frame while not modeling the camera motion at all.
These approaches cannot handle moving cameras as they typically require static camera views.
Static cameras have significant limitations, specifically for capturing complex motions and interactions where variations in perspective and occlusion are frequent.
Without the camera motion, estimations from a fixed viewpoint often lead to inaccurate poses, particularly in dynamic scenes or when the subject is partially occluded.
In contrast, moving cameras enable more accurate human pose estimation by providing additional spatial information compared to static cameras, even in cases where the subject is partially occluded from a static view.
By accounting for camera motion, we can better capture complex human-scene interactions and ensure accurate human pose reconstruction.
%
% Therefore, to advance these fields further, it is essential that robust models capable of handling the challenges posed by variation in depth and perspective are developed.
%
%%%%%
%
\par 
Compared to the amount of work that is done within root-relative coordinates, the area of 3D human reconstruction and tracking using dynamic cameras in the global coordinate system has seen far less progress~\cite{Yuan2021GLAMRGO,Ye2023DecouplingHA, Kocabas2023PACEHA}. 
One reason for this gap is the lack of comprehensive datasets that include, both, human and camera motion in the world frame along with respective ground truth annotations. 
Furthermore, even among the few datasets that exist~\cite{Zhang2021EgoBodyHB,Kaufmann2023EMDBTE}, none include the accompanying scene geometry together with the aforementioned ground-truth data. 
As a result, when methods are evaluated on these limited datasets, they may appear to produce plausible human motion. 
However, resulting motions often completely ignore scene geometry leading to artifacts such as human-environment intersection or unrealistic elevation from the scene.
Therefore, it is crucial to include ground-truth scene data as an integral part of the evaluation benchmark.
%
%%%%%
%
\par 
To truly assess whether a method can handle complex scenes and still recover precise human motion trajectories in the world frame, we collect an evaluation benchmark, MoviCam, which provides all necessary components: 
3D human model parameters in SMPL~\cite{Loper2023SMPLAS} format, human motion trajectory in the world frame, dynamic camera trajectory, the scene geometry, and human-scene contact labels. 
Sequences are captured in a complex scene, featuring objects like a table, a step stool, and various carpets scattered across the floor, where a person, for example, interacts with the environment by climbing the step. 
In this way, we capture not only human motion but also their interaction with the scene.
Together, these elements provide a holistic and consistent 3D understanding of the physical world and human interactions within it. 
Using this evaluation benchmark, we comprehensively evaluate the performance of recent methods and demonstrate the current limitations of the state of the art. 
%
%%%%%
%
% \par 
% %
% Integrating raw data from an external dynamic camera with the data from the capture studio poses several challenges.
% %
% The first challenge is obtaining the camera extrinsics for the dynamic camera. 
% %
% To address this, we design a system where the dynamic camera is mounted to a checkerboard, enabling accurate camera tracking with the multi-view capturing system. 
% %
% The setup is illustrated in Figure~\ref{fig:setup}. 
% %
% For the synchronization of the setup, a fixed FPS is set and soft-synchronization is performed.
% %
% \input{fig/setup} 
% %
% \CL{should the data capture details be included in the intro or the dataset section}
%
%%%%%
%
\par 
To handle moving cameras and non-flat terrains, we propose a physics-based method to optimize human pose in a way that is plausible with respect to the scene, physical laws, and the camera motion. 
First, 4DHumans~\cite{Goel2023HumansI4}, a state-of-the-art human pose estimator, is employed to obtain the subject’s pose. 
In parallel, DROID-SLAM~\cite{Teed2021DROIDSLAMDV} estimates the moving camera's trajectory. 
Then, the subject is positioned in global coordinates using the estimated camera trajectory.
At this stage, the subject's pose often exhibits jitters and unrealistic interactions with the scene such as penetration.
Hence, in the next step, the estimated pose and translation are refined with a physics optimizer module, enhancing the one in PIP~\cite{Yi2022PhysicalIP}. 
The scene geometry is also incorporated into the proposed physics module, making it scene-aware. 
With the information from the scene geometry, it is ensured that the subject interacts with objects in the environment appropriately, avoiding implausible collisions, and responds naturally to their surroundings based on the given video input.
\textit{Importantly, none of the previous methods individually solve the problem of physics-based human pose estimation in non-flat terrains.}
%
%%%%%
%
\par 
Our contributions can be summarized as follows: 
\begin{itemize}
    \item  We introduce a novel evaluation benchmark, \textbf{MoviCam}, for human motion tracking with a moving camera in a complex scene. 
    To the best of our knowledge, it is the first dataset to include detailed scene geometry along with global human motion and moving camera trajectories, providing accurate 3D human pose and shape, and human-scene contact labels.
    \item We propose a physics-based method, \textbf{PhysDynPose}, which combines a state-of-the-art human pose estimator with a scene-aware, physics-based motion optimizer. 
    \item We highlight where current state-of-the-art methods fail in the proposed benchmark, identifying key challenges for future improvement.
\end{itemize}

%%%%%%%%%%%%%%%%%%%%%%%%
%
\section{Related Work}
\label{sec:related_work}

Human motion capture is an active research area with many studies. 
Since our method focuses on proposing an evaluation benchmark for global human motion and camera trajectory, along with a physics-based human pose optimizer model, we only discuss previous motion capture datasets, monocular global human trajectory estimation methods, and physics-based models for physically plausible human pose recovery.
We do not discuss the extensive body of work that focuses on root-relative pose estimation using keypoints~\cite{Akhter2015PoseconditionedJA, Mehta2016Monocular3H, Mehta2017SingleShotM3} and joint angles~\cite{Pavlakos2019ExpressiveBC, Kolotouros2019LearningTR, Xu2020GHUMG, Li2020HybrIKAH}.

%-------------------------------------------------------------------------
\subsection{Motion Capture Datasets}
Most motion capture datasets rely on static cameras~\cite{rich, Ionescu2014Human36MLS, Zhang20204DAG, Yu2018HUMBIAL, Patel2021AGORAAI, Mehta2016Monocular3H, Hassan2019Resolving3H, Marcard2016HumanPE, Yi2022HumanAwareOP, Marcard_2018_ECCV, Mahmood2019AMASSAO, Joo2018TotalCA, Sigal2010HumanEvaSV, geo_aff}.
Human3.6M~\cite{Ionescu2014Human36MLS}, HumanEva~\cite{Sigal2010HumanEvaSV}, TotalCapture~\cite{Joo2018TotalCA}, and AMASS~\cite{Mahmood2019AMASSAO} use optical markers to capture high-quality motion but are limited to controlled studio settings with static cameras. GPA~\cite{geo_aff}  and RICH \cite{rich} include scene geometry but lacks dynamic camera motion.
We introduce a dynamic camera alongside multi-view static cameras, capturing more complex scenes. Unlike most datasets, we provide global human and camera trajectories, enabling better motion analysis.
Since few datasets exist for dynamic camera settings, some works create synthetic ones. GLAMR~\cite{Yuan2021GLAMRGO} simulates moving cameras via image cropping, while TRACE~\cite{Sun2023TRACE5T} synthesizes dynamic viewpoints from static and panoramic videos. However, these lack real-world perspective effects.
3DPW~\cite{Marcard_2018_ECCV} records dynamic camera motion in real-world settings. HPS~\cite{Guzov2021HumanPS} and EgoBody~\cite{Zhang2021EgoBodyHB} provide egocentric views with registered SMPL poses. EMDB~\cite{Kaufmann2023EMDBTE} includes global human and camera trajectories but lacks detailed scene information.
SLOPER4D~\cite{sloper4d} captures large-scale urban human motion with 3D poses, global camera trajectories, and LiDAR-based scene data but lacks accurate foot-ground contact. 
Our MoviCam dataset improves on this by providing scene geometry and foot-scene contact, essential for precise full-body motion estimation.
See Table \ref{table:comparison} for a detailed comparison.
\begin{table*}[ht!]
    \centering
    \begin{tabular}{l|c|c|c|c|c|c|c} 
        \hline
        & \textbf{Number of} & \textbf{Number of} & \textbf{Global} & \textbf{Camera} &  &  & \textbf{Contact} \\
        \textbf{Dataset}  & \textbf{Frames} & \textbf{Sequences} & \textbf{Motion} & \textbf{Trajectory} & \textbf{3D-Scene}  & \textbf{Real} & \textbf{Information} \\ 
        \hline
        3DPW~\cite{Marcard_2018_ECCV}      
        & 51k & 7
        & \textcolor{purple}{$\times$}   & \textcolor{purple}{$\times$} 
        & \textcolor{purple}{$\times$}  & \textcolor{teal}{$\checkmark$} 
        & \textcolor{purple}{$\times$}  \\
        EgoBody~\cite{Zhang2021EgoBodyHB}    
        & 220k & 125       
        & \textcolor{teal}{$\checkmark$} & \textcolor{teal}{$\checkmark$} 
        & \#        & \textcolor{teal}{$\checkmark$} 
        & \textcolor{purple}{$\times$} \\
        Dynamic Human3.6M~\cite{Yuan2021GLAMRGO} 
        & 51k & 7               
        & \textcolor{teal}{$\checkmark$} & $\ast$ 
        & \textcolor{purple}{$\times$} & \textcolor{purple}{$\times$}  
        & \textcolor{purple}{$\times$} \\
        DynaCam~\cite{Sun2023TRACE5T}    
        & 48k & 500        
        & \textcolor{teal}{$\checkmark$} & $\ast$        
        & \textcolor{purple}{$\times$} & \textcolor{purple}{$\times$}  
        & \textcolor{purple}{$\times$} \\
        EMDB~\cite{Kaufmann2023EMDBTE} 
        & 104k & 81  
        & \textcolor{teal}{$\checkmark$} & \textcolor{teal}{$\checkmark$}         
        & \textcolor{purple}{$\times$} & \textcolor{teal}{$\checkmark$} 
        & \textcolor{purple}{$\times$} \\
        SLOPER4D\cite{sloper4d}             
        & 100k & 15 $\dagger$
        & \textcolor{teal}{$\checkmark$} & \textcolor{teal}{$\checkmark$} 
        & \textcolor{teal}{$\ddagger$} & \textcolor{teal}{$\checkmark$} 
        & \textcolor{purple}{$\times$} \\
        \hline
        Ours            
        & 22k & 7 
        & \textcolor{teal}{$\checkmark$} & \textcolor{teal}{$\checkmark$} 
        & \textcolor{teal}{$\checkmark$} & \textcolor{teal}{$\checkmark$} 
        & \textcolor{teal}{$\checkmark$} \\
        \hline
    \end{tabular}
    \caption{Comparison of dynamic camera datasets with different features. * represents simulated camera movement and \# refers to an incomplete scene mesh. $\dagger$ refers to the whole dataset, but only $6$ sequences are publicly released. $\ddagger$ refers to included scene geometry but it is not explicitly optimized. Note that ours is the only dataset that also provides the optimized 3D scene and foot contact information in addition to being captured from a moving camera, making it a suitable test setting for human-scene interaction works.}
    \label{table:comparison}
\end{table*}
%
%Most current motion capture datasets provide videos captured with static cameras, rather than dynamic ones, for learning or evaluating human motion~\cite{Ionescu2014Human36MLS, Zhang20204DAG, Yu2018HUMBIAL, Patel2021AGORAAI, Mehta2016Monocular3H, Hassan2019Resolving3H, Marcard2016HumanPE, Yi2022HumanAwareOP, Marcard_2018_ECCV, Mahmood2019AMASSAO, Joo2018TotalCA, Sigal2010HumanEvaSV, geo_aff}. 
%
%Among these works, Human3.6M~\cite{Ionescu2014Human36MLS}, HumanEva~\cite{Sigal2010HumanEvaSV}, TotalCapture~\cite{Joo2018TotalCA}, AMASS~\cite{Mahmood2019AMASSAO} use optical markers in their motion capture systems to collect high-quality, large-scale datasets. 
%
%While they offer precise annotations, their studio setups are limited to static cameras, and the captured data feature low-variance background scenes. 
%
%GPA~\cite{geo_aff} provides global human pose and complex scene geometry, but it does not have any sequences with dynamic cameras.
%
%Like these datasets, we are limited to a studio setup, but we use a dynamic camera alongside multi-view static cameras, and include scenes with varying objects to increase scene complexity. 
% 
%Moreover, most of these datasets do not provide global human trajectory, whereas we provide both global human and global camera trajectories of our moving camera.
%
\par
%
%Since there are not many suitable training and evaluation datasets for global human motion estimation in a dynamic camera setting, some works propose their own datasets to evaluate their methods. 
%
%Dynamic Human3.6M is a synthetic dataset proposed by GLAMR~\cite{Yuan2021GLAMRGO}, simulating moving cameras using image-cropping methods. 
%
%DynaCam is another synthetic dataset introduced by TRACE~\cite{Sun2023TRACE5T} by simulating moving cameras in in-the-wild panoramic videos, as well as videos captured by static cameras. 
%
%However, these synthetic datasets do not fully reflect the complex nature of dynamic cameras in real world, as they are synthesized from static viewpoint videos and lack perspective effects. 
%
\par

\subsection{Monocular Global 3D Human Trajectory Estimation}
Recovering human motion in the world frame from a monocular dynamic camera is challenging.
GLAMR~\cite{Yuan2021GLAMRGO} separates human and camera motion using learned motion priors.
SLAHMR~\cite{Ye2023DecouplingHA} jointly optimizes human and camera motion to resolve scene scale ambiguity.
PACE~\cite{Kocabas2023PACEHA} aligns motion with scene features and human pose.
TRACE~\cite{Sun2023TRACE5T} learns a 5D representation (space, time, identity) for tracking human motion in global coordinates. However, these methods assume a flat floor and suffer from drift in long sequences.
GLAMR ignores scene context by cropping human poses. SLAHMR and PACE are computationally expensive due to motion priors. PACE and TRACE also use synthetic datasets without detailed scene geometry or ground-truth camera motion, limiting real-world tracking accuracy.
Unlike these methods, we optimize human motion on non-flat terrain using physics-based constraints and scene geometry. This enables accurate reconstruction of complex interactions like climbing stairs or navigating slopes while reducing trajectory drift.
WHAM~\cite{Shin2023WHAMRW} recovers human meshes and global positions but struggles with elevation shifts from jumping or squatting due to inaccurate foot-scene interactions.
BodySLAM~\cite{bodyslam} and TRAM~\cite{tram} estimate global trajectories using SLAM and learned priors but lack physics-based reasoning, leading to implausible motion.
Our method integrates physical constraints and scene geometry, ensuring more accurate motion recovery and preventing drift.
\subsection{Physics-based Methods for Human Pose Estimation}
Most existing methods for recovering human motion from dynamic monocular cameras rely on kinematic models, which represent motion through joint rotations and positions without considering physical forces. 
While these approaches directly capture 3D body geometry for training, they often produce unrealistic results, such as body-scene penetration or implausible interactions. These limitations become pronounced in dynamic environments where friction and gravity influence movement (e.g., jumping, squatting, climbing stairs).
To address these issues, some works~\cite{Shimada2020PhysCap, Shimada2021NeuralM3, Yuan2021SimPoESC, Li2022DDLH} incorporate physics-based constraints for more realistic motion.
PhysCap~\cite{Shimada2020PhysCap} optimizes human pose under physical constraints using PyBullet~\cite{coumans2016pybullet}. Neural PhysCap~\cite{Shimada2021NeuralM3} extends this by learning PD controller gains via a neural network.
PIP~\cite{Yi2022PhysicalIP} improves on PhysCap with two PD controllers (rotation and position) for real-time IMU-based motion capture.
D\&D~\cite{Li2022DDLH} introduces physics-based equations for motion recovery in moving camera settings. However, all these methods assume a flat ground and ignore scene geometry, limiting their applicability to real-world scenarios.
Our method, PhysDynPose, builds on PIP’s physics-aware motion optimizer but differs in inputs and processing. While PIP relies on sparse IMU-based inertia measurements, we use video of human-scene interactions. 
Instead of RNNs to estimate contacts and joint states, we leverage 4DHumans~\cite{Goel2023HumansI4} for human mesh recovery and tracking. By incorporating detailed scene geometry and physics-based constraints, our approach enables accurate reconstruction of human motion in complex environments, reducing trajectory drift and improving realism.
\section{Dataset}
\label{sec:dataset}
This section introduces our new evaluation benchmark MoviCam. 

\subsection{Dataset Capture}
Extracting reliable ground-truth for scene mesh and dynamic camera trajectory is challenging. 
Hence, to provide an evaluation dataset with highly precise ground-truth data for camera trajectories, scene mesh, human pose, and motion in the world frame, we captured our dataset in a controlled studio environment. 
%
% Long sequences, minimal background reflections, and rich features in the scene setup are essential for reliable scene reconstruction.
%
% All scenes were captured with multiple static cameras as well as a lightweight freely-moving camera.
%So, an external lightweight camera that could move freely within the setup is preferred, allowing us to capture longer sequences while preventing checkerboard deformation. 
%\CL{Rephrase: Therefore, an external light-weight camera that can move freely in the studio would be preferred for the dynamic capture} 
%

%%% CHECK THIS AGAIN BUT THIS IS A DIRE MEASURE NOW
% The scene setup with objects is illustrated in Figure \ref{fig:objects} together with the overall system setup.
% \input{fig/objects}

% \subsection{Data Collection and Annotation}

\par \noindent \textbf{Data Collection.}
Our studio setup featured 120 multi-view synchronized static cameras, capturing images up to 4K resolution for accurate scene reconstruction and human motion tracking.
Post-processing utilized 34 cameras at 2K resolution.
We employed Captury~\cite{thecaptury2020}, a markerless system ensuring non-intrusive, natural motion tracking for motion capture.
Additionally, a SONY RX10 sports camera (1080x1920) was used for dynamic capture, serving as input to our method and baselines.
%
% A checkerboard is attached to the dynamic camera and used in conjunction with the multi-view studio cameras to track the movement of the moving camera.
A checkerboard attached to the moving camera enabled its tracking via static cameras. 
Two individuals managed the capture: one interacting with the scene and another controlling the moving camera.
% Two people are required for the capturing process: one person as the subject interacting with the scene and another person to operate the moving camera, adjusting its position and distance while filming to ensure the stream remains steady and the subject is not fully occluded. 

\par \noindent \textbf{Hand-eye Calibration for Camera Trajectory.}
\label{sssection:handeye}
Recovering accurate trajectories for the dynamic camera is one of the key challenges in the data capture process.
Towards this end, we utilize the hand-eye calibration approach of Strobl et al.~\cite{Strobl2006OptimalHC} where we utilize the following transformations between: (a) the external camera and the floor checkerboard $(T_{EF})$, (b) the external camera and the head checkerboard $(T_{EH})$, (c) and the moving camera of the floor checkerboard $(T_{MF})$.  
With the following equation
\begin{equation}
\label{eq:calibration}
T_{\text{Hand-eye}} = T_{EH}^{-1} T_{EF} \cdot T_{MF}^{-1}
\end{equation} we can estimate the transformation between the head checkerboard and the moving camera $(T_{\text{Hand-eye}})$.

\subsection{Ground-truth Acquisition}
\label{ssection:gt_acquisition}
We obtain the ground-truth data from the raw streams captured by the moving camera and the multi-view cameras after recording the subject in our complex scene setup. We aim to provide:
\begin{itemize}
    \item Dense scene geometry as a mesh and height map
    \item Human pose and shape
    \item Human motion trajectory in the world frame
    \item Global camera trajectory
    \item Contact labels between human and scene
\end{itemize}

\noindent \textbf{Dense Scene Geometry as a Mesh and Height Map.} The ground-truth scene mesh was generated through a reconstruction process using multi-view images captured in 4K resolution from 120 cameras.
To obtain the height map, the scene mesh is loaded into Pybullet~\cite{coumans2016pybullet}, and a grid of the scene with resolution $1024 \times 1024$ is generated. 
We then shoot rays through every grid cell from above the scene to below, and record the height of the first point that intersects with the scene.
Finally, we obtain a height map $h$ that can be queried with foot joint positions $(x,z)$ as $h(x,z)$. 

\noindent \textbf{Human Pose, Shape and Motion Trajectory in World Frame.} 
Since the data is captured using the Captury~\cite{thecaptury2020} system, we obtain the skeleton pose and motion in the Captury skeleton format. 
However, most human motion tracking methods use SMPL~\cite{Loper2023SMPLAS}.
To bridge this gap, we align SMPL 3D joints with the Captury skeleton and attach markers at SMPL joint locations.
The shape parameter is estimated from the first 100 T-pose frames and then fixed.
Finally, we obtain pose and translation per frame by processing the full sequence using Captury tracking with SMPL joint markers as input.

\noindent \textbf{Global Camera Trajectory.}
The images from 34 studio cameras are used to triangulate the checkerboard position in each frame and the estimated poses from each camera are averaged.
%
% As previously mentioned in Section~\ref{sssection:handeye}, the checkerboard-to-camera transformation matrix is obtained through hand-eye calibration. 
%
We then apply the $(T_{\text{Hand-eye}})$ from Equation~\ref{eq:calibration} to the estimated checkerboard poses to get the corresponding moving camera poses. 
%
% However, narrow angles of the checkerboard in multi-view images may introduce outliers, causing the averaged pose to deviate and resulting in jittery projected keypoints.
% %
% To address this issue, we compute the median translation vector and filter estimations based on deviations from this median.
% %
% This approach effectively reduces outliers and smooths the camera poses.
% However, outliers may occur due to narrow angles of the checkerboard in the multi-view images. Because of these outliers, the averaged pose might deviate from the actual position, causing the projected keypoints onto the images to appear jittery. To address this issue, the median translation vector is computed, and the estimations are filtered using a threshold based on the deviation from the median translation vector. This technique resolves the outliers and jitter issues in the camera poses.

\noindent \textbf{Contact Labels Between Human and Scene.} 
After obtaining the precise scene mesh and human pose, ground-truth contact labels are generated by calculating the distance between the foot joint location and the closest scene vertex, following~\cite{Shimada2020PhysCap}. 
If the computed distance is less than $5$ cm, it is labeled as "in contact"; otherwise, it is labeled as "not in contact". 

\subsection{Dataset Overview}
\begin{figure}[t]
  \centering
  \includegraphics[width=.99\linewidth]{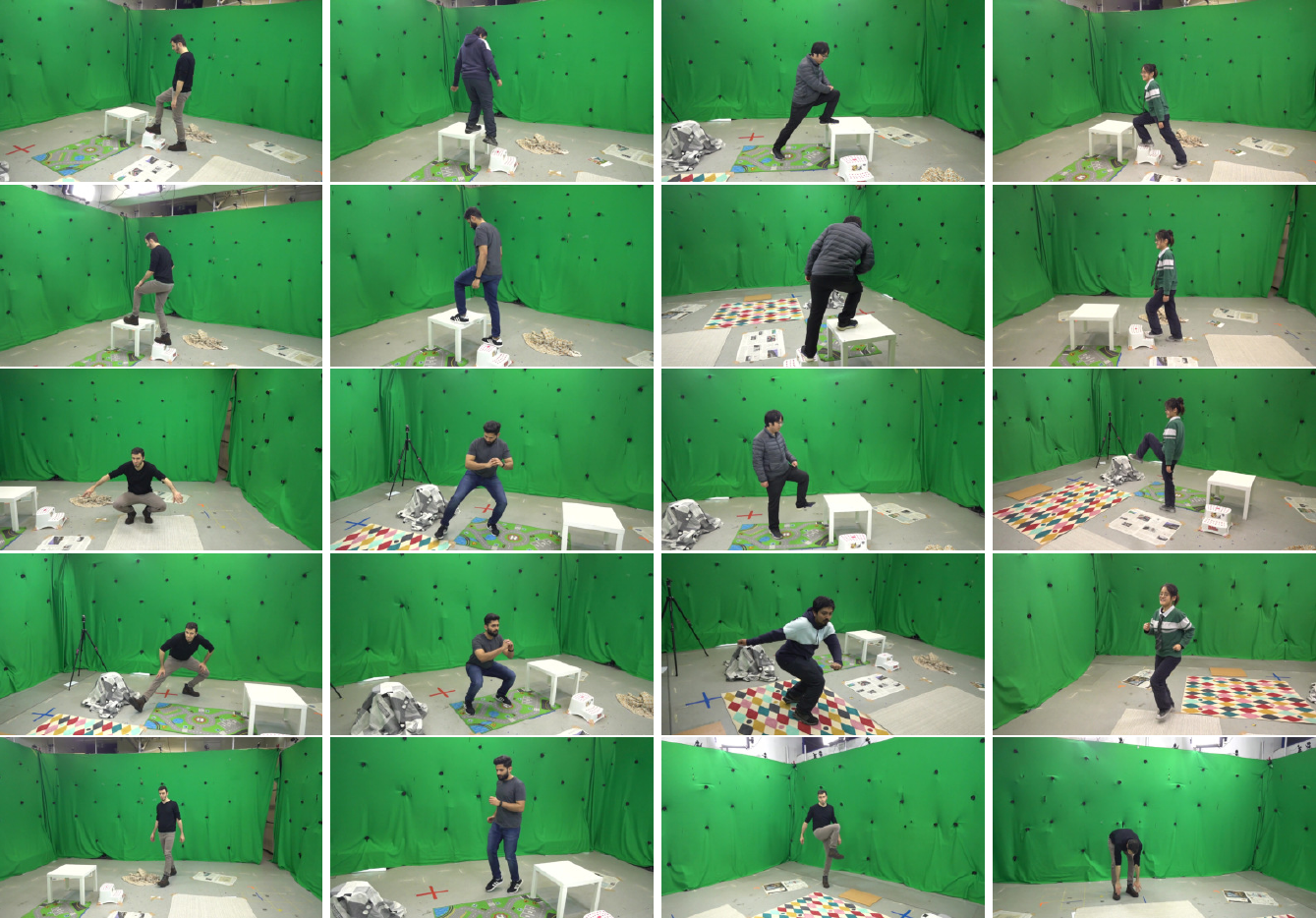}
  \caption{\label{fig:interaction}
           Example interactions in our proposed MoviCam dataset.}
\end{figure}

The dataset consists of $7$ sequences: $5$ sequences on non-flat ground and $2$ sequences on a flat surface. 
Each sequence features a different individual interacting with the scene, captured with different moving camera trajectories. 
These interactions range from walking and jumping to stretching and squatting (see Figure \ref{fig:interaction}).
With $7$  participants of varying heights and weights, the dataset contains approximately $22{,}000$ images. 
%
% Sequence durations are listed in Table \ref{tab:nameduration}.
% \input{tab/name_duration}

% The \textit{non-flat ground} setup includes diverse objects and patterned carpets, while the \textit{flat ground} setup features only a flat floor. 
%
We provide a scene mesh for the non-flat ground setup.
Overall, our dataset includes dense scene geometry, ground-truth 3D human pose and shape in SMPL format (24 joints, 300 shape parameters), global camera poses (extrinsics), and contact labels for the left/right toes and heels.

\section{Method}
\label{sec:method}

\begin{figure*}[t]
  \centering
  \includegraphics[width=.89\linewidth]{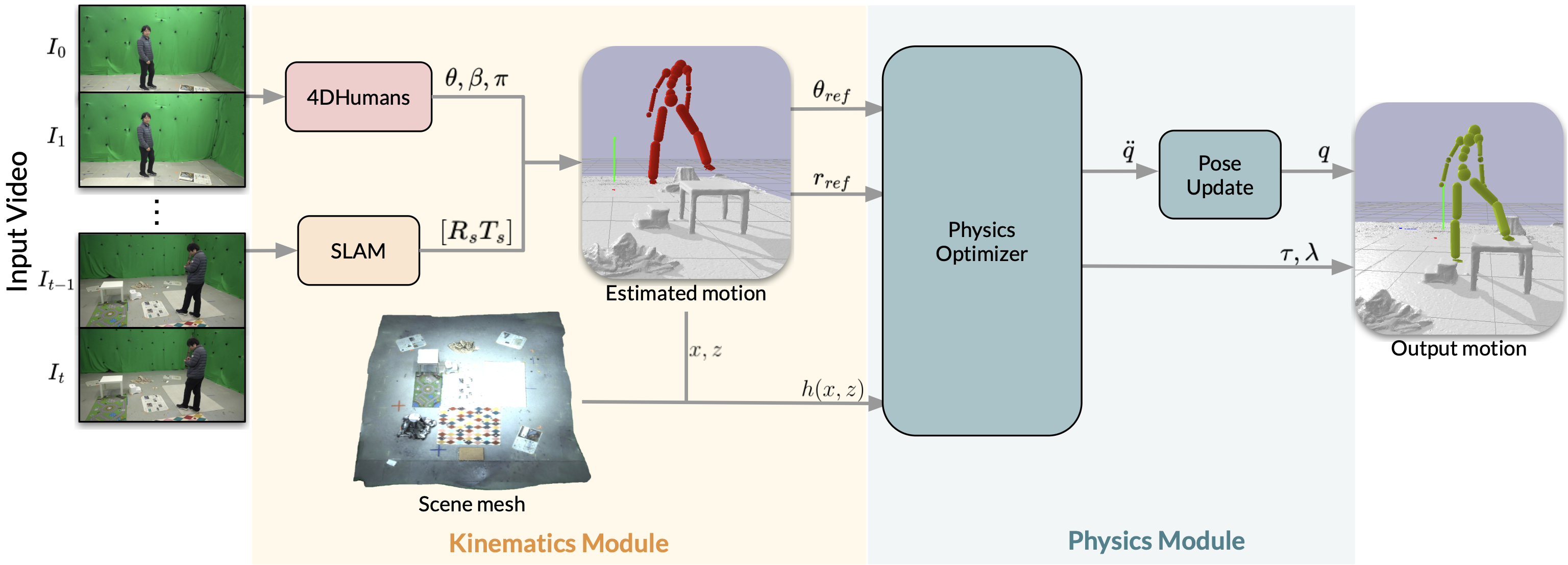}
  \caption{\label{fig:pipeline}
           \textbf{Overview of PhysDynPose}. We first use 4DHumans~\cite{Goel2023HumansI4} to estimate human motion in a root-relative frame and employ DROID-SLAM~\cite{Teed2021DROIDSLAMDV} to capture the dynamic camera trajectory. Next, a physics- and scene-aware motion optimizer refines the estimated motion. This process produces physically plausible human motion, along with joint torques and ground reaction forces.}
\end{figure*}

Previous methods~\cite{Yuan2021GLAMRGO,Ye2023DecouplingHA,Kocabas2023PACEHA,Shin2023WHAMRW} have demonstrated high accuracy in recovering human pose and shape, along with robust global tracking capabilities. 
However, in scenarios where the human interacts with a non-flat scene, current methods are prone to producing physically implausible results. 
We propose PhysDynPose, which integrates scene geometry and physical constraints to produce coherent global motion in complex environments.
An overview of our method is shown in Figure ~\ref{fig:pipeline}.
The inputs to our method are an image sequence $\mathbf{I} = \{\mathbf{I}_t\}_{t=1}^{T}$ with $T$ frames capturing a person navigating through non-flat terrain, scene mesh and foot contact labels.
For each input frame $\mathbf{I}_t$, our method outputs the subject's pose, $\pose$, in terms of joint angles $\boldsymbol{\theta}$ and root translation $\jposition_{\text{root}}$ in world frame, following the SMPL~\cite{SMPL:2015} body model. Additionally, physical properties related to the subject, such as ground reaction forces $\grf$ and joint torques $\torque$, are computed.
%
% For joint angles, $\theta$, we follow the SMPL kinematic tree with $J=24$ joints. The pose parameter has the shape $\jangle \in \reel^{J \times 3 \times 3}$. For the shape parameter, the first $10$ principal components are used following 4DHumans so that it has the shape $\shape \in \reel^{10}$. The SMPL model outputs joint positions $\jposition \in \reel^{3J}$. Note that the pose parameter $\jangle$ consists of both the global orientation $\jangle_g \in \reel^{1 \times 3 \times 3}$ and the body pose $\jangle_b \in \reel^{23 \times 3 \times 3}$.
%
Overall, the per-frame output of our method is $(\jposition_{\text{root}}, \boldsymbol{\theta}, \grf, \torque)$.
We use a plug-and-play approach for our models which do not require any additional training. 
In our dataset and method, we assume y-axis is up.
\par
% We utilize two previous works in our method as two modules: 4DHumans and Physical Inertial Poser (PIP). \RD{Never write such a line :-).}
As illustrated in Figure~\ref{fig:pipeline}, our method follows a 2-stage pipeline.
In the first stage (kinematics module), we first estimate the 3D body pose and camera trajectories in the world frame.
In the second stage (physics module), these estimates drive a dual PD controller, whose outputs are refined through a quadratic optimization routine.
The optimized joint accelerations are used to update the pose of the character in simulation.

\subsection{Kinematics Module}
The goal of the kinematics module is to provide the initial estimates of the human pose, shape and camera translation for every frame of the input video. 
To this end, we employ the state-of-the-art methods for monocular motion capture and camera trajectory estimation.
We build on 4DHumans~\cite{Goel2023HumansI4} for human motion capture and use DROID-SLAM~\cite{Teed2021DROIDSLAMDV} for dynamic camera trajectory estimation.
Due to SLAM suffering from scale ambiguity, we align it using the first two frames of the ground-truth camera trajectory.
Since 4DHumans predicts global orientation $\jangle_g$ and camera translation $\camera$ in a root-relative frame, we convert its estimates to world coordinate system using the estimated camera trajectories from DROID-SLAM.
We follow 
\begin{equation}
\begin{aligned}
{}^w\jangle_g &= \mathbf{R_{S}}^{-1} {}^c\jangle_g, \quad \\
{}^w\camera &= \mathbf{R_{S}}^{-1} \left( {}^c\camera - \mathbf{T_{S}} \right),
\end{aligned}
\end{equation} where ${}^w\jangle_g$ is to the root orientation in world frame, ${}^c\jangle_g$ is the root orientation in camera frame, ${}^w\camera$ is the global root translation in world frame, and ${}^c\camera$ is the root translation in camera frame,  $\mathbf{R_{S}}$ is the estimated camera rotation, $\mathbf{T_{S}}$ is the estimated camera translation.
To reduce jitter, we apply a One-Euro-Filter \cite{Casiez20121F} with a minimum cut-off frequency of $0.004$ and a speed coefficient of $0.7$ to the pose and translation in world frame.
Finally, we recover the estimated human pose $\jangle_{ref}$ and joint positions $r_{ref}$ from 4DHumans in world coordinates, to be refined in the following physics module.

\subsection{Physics Module}
The physics module refines initial kinematic estimates to address artifacts like jitter and scene penetration (see Figure~\ref{fig:pybullet_qualitative}, rows 2-3).
%
% The estimations from the kinematics module may still exhibit artifacts such as jitter or undesired interactions with the environment, including ground and scene penetration, like a human walking through a table in the scene, as illustrated in the last row of Figure~\ref{fig:pybullet_qualitative}.
%
To address these issues, inspired by~\cite{Yi2022PhysicalIP}, we extend its physics-aware motion optimizer to explicitly incorporate scene awareness via scene-height map and $\jposition_{\text{root}}$ supervision.
%, using transformed 4DHumans estimates instead of IMU data.
%
The input to the physics module consists of the estimated pose $\jangle_{ref}$ and joint positions $r_{ref}$ in world frame.
These inputs serve as the reference in the physics optimizer. 
Given these estimates, the physics optimizer calculates physically plausible accelerations, updating the character’s pose $\pose = [\jposition_{\text{root}} \jangle]$ frame-by-frame within PyBullet \cite{coumans2016pybullet} using a floating-base humanoid. 
The humanoid character's position is initialized according to the estimated root joint position $r_{root}$ in world frame.

In the optimizer, the character's pose is updated as
\begin{equation}
\label{eq:dynamic_updates_root}
\begin{aligned}
    \pose_{:3}^{(t+1)} &= \pose_{:3}^{(t)} + \dot{\pose}_{:3}^{(t)} \Delta t , \\ 
    \dot{\pose}_{:3}^{(t+1)} &= \dot{\pose}_{:3}^{(t)} + \ddot{\pose}_{:3}^{(t)} \Delta t ,
\end{aligned}
\end{equation} where $\dot{\pose}$ is the generalized velocity and $\ddot{\pose}$ is the generalized acceleration, and specifically, $\ddot{\pose}_{:3}$ is the root acceleration, $\dot{\pose}_{:3}$ is the root velocity and $\pose_{:3}$ is the root translation.

\subsubsection{Enhanced Physics Model}%Enhanced Physics Model}
\label{sssection:ephysicsmodel} %ephysicsmodel
We employ PIP's physics-based optimizer that has dual PD controllers with the same input-output structure, now including scene geometry and contacts as an additional inputs. Overall, we introduce two key enhancements:
\begin{itemize}
\item \textbf{Scene Geometry Integration}. 
For the friction cone and sliding constraints in the physics optimizer, PIP checks for scene penetration by evaluating the foot’s elevation and contact labels. 
Previously, this was determined as $r_{\text{foot}, y} < 0$ where $r_{\text{foot}, y}$ is the contacting foot joint's position in the y-axis, assuming a flat scene.
Instead, we integrate the height map $h(x,z)$ obtained from the scene mesh (Section~\ref{ssection:gt_acquisition}), leading to a more accurate penetration check as $r_{\text{foot}, y} < h(r_{\text{foot}, x}, r_{\text{foot}, z})$ where $r_{\text{foot}, x}$ and $r_{\text{foot}, z}$ are the contacting foot joint's positions along the x and z axes, respectively.
\item \textbf{Root Supervision}. 
We use the motion update rules in Equation~\ref{eq:dynamic_updates_root} to obtain 
\begin{equation}
\label{eq:root_supervision}
\begin{aligned}
\ddot{\pose}_{:3}^{(t)} &= \frac{1}{\Delta t^2} \left( \pose_{:3}^{(t+2)} -\pose_{:3}^{(t+1)} - \dot{\pose}_{:3}^{(t)} \Delta t \right)
\end{aligned}
\end{equation}  where $t$ represents the current frame and $\Delta t$ is set to $1/60$ second as PIP's system runs at 60 fps.
By supervising the root joint using future frames, we prevent long-sequence drift.
Since the humanoid's pose $\pose$ is initialized using the output estimates from the kinematic module, we know the initial states. 
Hence, we can perform these updates to help solve for the future states.
\end{itemize}

After our simple but effective additions to the physics optimizer in PIP, we have the following final optimization objective:
\begin{equation}
\label{eq:optimizer}
\begin{alignedat}{2}
\vspace{-5pt}
\arg \min_{\dot{\pose}, \lambda, \tau} & \quad \mathcal{E}_{\text{PD}} + \mathcal{E}_{\text{reg}} \\
\vspace{-5pt}
\text{s.t.} \quad & \tau + \jacobian_c^\top \lambda = \inertia \ddot{\pose} + \nonlinear & \quad & \text{(equation of motion)} \\
                  & \lambda \in \mathcal{F} & \quad & \text{(friction cone)} \\
                  & \dot{r_j}(\ddot{\pose}) \in \mathcal{C} & \quad & \text{(no sliding)} \\
                  & \ddot{\pose}_{:3}^{(t)} = \frac{\pose_{:3}^{(t+2)} - \pose_{:3}^{(t+1)}}{\Delta t^2} - \frac{\dot{\pose}_{:3}^{(t)}}{\Delta t} & \quad & \text{(no drifting)},
\end{alignedat}
\end{equation} where $\tau$ is the joint torques, $\inertia$ is the inertia matrix, $\nonlinear$ is the non-linear effects term, $\grf$ is the contact forces applied at contact points, $\jacobian_c$ is the contact point Jacobian matrix.
For further details on energy terms $\mathcal{E}_{\text{PD}}$ and $\mathcal{E}_{\text{reg}}$, and the friction cone $\mathcal{F}$ and no sliding constraints $\mathcal{C}$, refer to PIP~\cite{Yi2022PhysicalIP}.

\section{Experiments}
\label{sec:experiments}
State-of-the-art models and the proposed method are evaluated on our proposed evaluation benchmark MoviCam using global coordinates.
%
%
%-------------------------------------------------------------------------
\subsection{Metrics}
We evaluate the performance of the methods in two parts: (a) 3D reconstruction errors and (b) physical plausibility. 

\noindent\textbf{3D Reconstruction Errors.} 
%To evaluate the accuracy of 3D human pose and trajectory estimation, we first compute Mean Per Joint Position Error (MPJPE) and Procrustes-aligned MPJPE (PA-MPJPE) in $mm$. Then, highlighting the errors in world frame, W-MPJPE metric is reported in $mm$, which is MPJPE after the first frames of the predictions and ground-truth data are aligned, while WA-MPJPE is reported in $mm$, which is MPJPE metric after the entire trajectories of the predictions and ground-truth data are aligned. Additionally, following \cite{Shin2023WHAMRW}, we report Root Translation Error (RTE) in $\%$, normalized by the actual displacement of the subject. This metric is computed over the entire trajectory after rigid alignment.
To evaluate 3D human pose and trajectory estimation accuracy, we compute Mean Per Joint Position Error (MPJPE) and Procrustes-aligned MPJPE (PA-MPJPE) in mm. We also report W-MPJPE, which is MPJPE after aligning the initial frames of predictions and ground-truth data, and WA-MPJPE, which is after aligning all trajectories. Additionally, we follow \cite{Shin2023WHAMRW} in reporting Root Translation Error (RTE) as $\%$, normalized by the subject's actual displacement, calculated over the entire trajectory after rigid alignment.

\noindent \textbf{Physical Plausibility Metrics.}
%The objective of the physical plausibility metrics is to demonstrate how physically accurate the reconstructed human motion is and how plausible it is with respect to the scene. Therefore, we introduce three new physical plausibility metrics: $\%$ of frames with scene penetration, average penetration per frame in $mm$, and average distance above the scene in $mm$. $\%$ of frames with scene penetration shows the percentage of frames where the subject's feet penetrates the scene, including the objects within the scene. Average penetration per frame is the amount of penetration depth the subject's feet have with the scene in a frame on average. Average distance above the scene demonstrates how elevated the subject is from the scene in a frame on average. We utilize the ground-truth scene geometry and height map to compute these metrics.
%Finally, following \cite{Shimada2020PhysCap}, jitter is adapted as the temporal smoothness error in 3D coordinates in $mm/s$. Additionally, following \cite{Shin2023WHAMRW}, foot sliding, which is the averaged displacement of foot toe joints during contact, is implemented in $mm$. 
Physical plausibility metrics assess the accuracy of reconstructed motion relative to the scene. We introduce three new metrics: (1) $\%$ of frames with scene penetration, (2) average penetration depth per frame (mm), and (3) average height above the scene (mm), all computed using ground-truth scene geometry and height maps.

Following \cite{Shimada2020PhysCap}, we measure jitter as temporal smoothness error (mm/s). Additionally, based on \cite{Shin2023WHAMRW}, we compute foot sliding as the average toe joint displacement during contact (mm).
%-------------------------------------------------------------------------
\subsection{Competing Methods}
\label{ssection:baselines}
%We evaluate the widely used method GLAMR~\cite{Yuan2021GLAMRGO}, one of the state-of-the-art methods for human global trajectory WHAM~\cite{Shin2023WHAMRW}, current state-of-the-art for human pose and shape estimation 4DHumans~\cite{Goel2023HumansI4} on our proposed benchmark, and compare them to the proposed method. 
%
%For a fair comparison, we use the ground-truth orientation and translation of the first two frames of the sequences to initialize the position of the subject in GLAMR and WHAM. 
%
%To demonstrate how the proposed physics module improves the kinematic module, we also evaluate 4DHumans by converting its results in root-relative frame to world frame using the estimated moving camera extrinsics that we obtained by DROID-SLAM.
We evaluate GLAMR~\cite{Yuan2021GLAMRGO}, WHAM~\cite{Shin2023WHAMRW}, and 4DHumans~\cite{Goel2023HumansI4} on our benchmark and compare them to our method. For fairness, we initialize GLAMR and WHAM using ground-truth orientation and translation from the first two frames. 
To demonstrate the impact of our physics module, we also evaluate 4DHumans by transforming its root-relative results into the world frame using estimated camera extrinsics from DROID-SLAM.

%-------------------------------------------------------------------------
\subsection{Results and Comparison}
\label{ssec:results}
\begin{figure}[t]
  \centering
  \includegraphics[width=.99\linewidth]{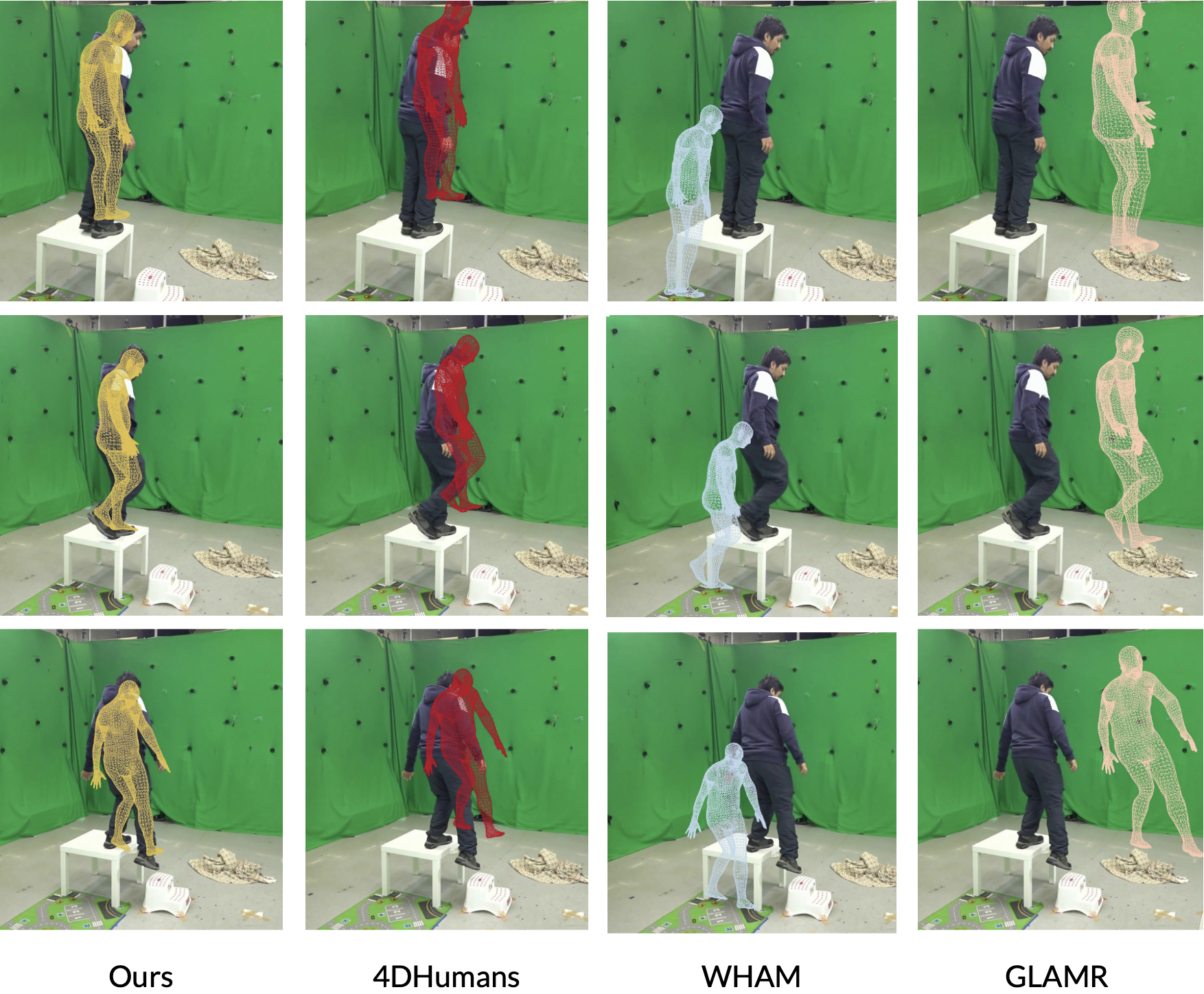}
  \caption{\label{fig:overlay_qualitative}
           Qualitative comparison between our method and previous methods. Each row presents a different frame from sequence 4, showing the person interacting with non-flat ground. We visualize human motion estimation results as meshes and project them back onto the input frames, overlaying the reconstruction result and the corresponding frame. Note that our results overlay to the input frames more accurately compared to the previous methods.}
\end{figure}

\begin{figure}[t]
  \centering
  \includegraphics[width=.99\linewidth]{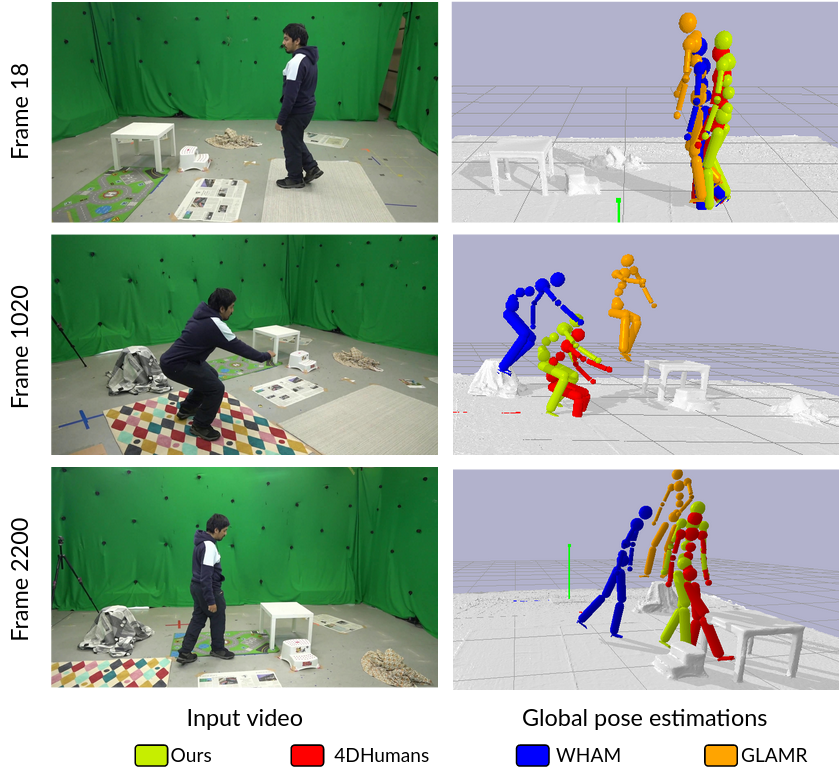}
  \caption{\label{fig:pybullet_qualitative}
           Qualitative comparison of the methods visualized in Pybullet for several frames of sequence 4. Note that even though all the estimated motions start from approximately the same point, as the sequence progresses, the competing methods suffer from drift and inaccurate elevation from the ground. Additionally, 4DHumans penetrates the scene as observed in second and third rows. In contrast, our method results in more accurate global trajectory and physically plausible pose with respect to the scene. }
\end{figure}
Tables \ref{table:recon} and \ref{table:physics} compare our method with state-of-the-art models, averaging metrics across sequences in flat and non-flat scenes. Figures \ref{fig:overlay_qualitative} and \ref{fig:pybullet_qualitative} show qualitative results.
4DHumans achieves the lowest MPJPE for motion reconstruction, while WHAM performs best at PA-MPJPE. Our method excels in trajectory estimation, outperforming others in W-MPJPE and RTE, while its WA-MPJPE is close to 4DHumans. Note that PA-MPJPE does not always correlate with physically plausible results, as seen in Figure~\ref{fig:pybullet_qualitative} where WHAM (blue) is the model with the best PA-MPJPE.
%Tables \ref{table:recon} and \ref{table:physics} offer a detailed quantitative comparison of our method against state-of-the-art models, with performance metrics evaluated as weighted averages across multiple sequences in two scenes, non-flat and flat. This evaluation protocol ensures that the results reflect each model's overall capability, accounting for variations in sequence length and complexity. Additionally, Figures \ref{fig:overlay_qualitative} and \ref{fig:pybullet_qualitative} shows our qualitative results on our benchmark. 
%In terms of motion reconstruction, 4DHumans leads in accuracy at MPJPE while WHAM performs the best at PA-MPJPE. For evaluating the accuracy of trajectory estimation, which is demonstrated by W-MPJPE and RTE, we outperform other models. For WA-MPJPE, we are quite close to 4DHumans.
% In terms of jitter, our method performs worse than WHAM and 4DHumans. This is due to inaccurate root position coming from SLAM-estimated camera trajectory. When the camera extrinsics are inaccurate, the root position of the subject is consequently inaccurate, resulting in the subject standing at a wrong position in the scene. In our method, this results in subject's wrong interaction with the scene, causing jitters in the motion. For instance, if the subject goes near the table in the scene without actually interacting with it, due to the estimated camera trajectory, the root position of the subject can appear up and down on the table in consecutive frames, which causes high jitters. 

\begin{table*}[ht!]
\centering
\begin{tabular}{c|c|c|c|c|c|c}
\hline
\textbf{Scenes} & \textbf{Models} & \textbf{MPJPE} $\downarrow$ & \textbf{PA-MPJPE} $\downarrow$ & \textbf{W-MPJPE} $\downarrow$ & \textbf{WA-MPJPE} $\downarrow$ & \textbf{RTE} $\downarrow$ \\
\hline
\multirow{4}{*}{\rotatebox[origin=c]{90}{\textbf{Non-flat}}}
& GLAMR \cite{Yuan2021GLAMRGO}      & 236.26 & \cellcolor{orange!30}46.62 & 1968.29 & 1013.18 & 2.92 \\
& WHAM \cite{Shin2023WHAMRW}        & 189.62 & \cellcolor{red!20}33.88 & 1352.06 & 698.58 & 2.19 \\
& 4DHumans \cite{Goel2023HumansI4}  & \cellcolor{red!20}128.01 & 51.75 & \cellcolor{orange!30}833.57 & \cellcolor{red!20}417.83 & \cellcolor{orange!30}1.18 \\
& \textbf{Ours}                     & \cellcolor{orange!30}162.09 & 64.11 & \cellcolor{red!20}779.60 & \cellcolor{orange!30}418.65 & \cellcolor{red!20}1.16 \\
% & GT                                & 0.0 & 0.0 & 0.0 & 0.0 & 0.0 \\
\hline
\multirow{4}{*}{\rotatebox[origin=c]{90}{\textbf{Flat}}} 
& GLAMR \cite{Yuan2021GLAMRGO}      & 199.85 & 45.86 & 3680.37 & 1521.66 & 6.68  \\
& WHAM \cite{Shin2023WHAMRW}        & 249.51 & \cellcolor{red!20}33.93 & 3220.94 & 838.06 &  4.16  \\
& 4DHumans \cite{Goel2023HumansI4}  & \cellcolor{red!20}105.70 & \cellcolor{orange!30}44.84 & \cellcolor{orange!30}1185.14 & \cellcolor{orange!30}500.93 & \cellcolor{orange!30}1.93  \\
& \textbf{Ours}                     & \cellcolor{orange!30}134.58 & 58.95 & \cellcolor{red!20}1092.72 & \cellcolor{red!20}489.72 & \cellcolor{red!20}1.87  \\
% & GT                                & 0.0 & 0.0 & 0.0 & 0.0 & 0.0 \\
\hline
\end{tabular}
\caption{Evaluating the 3D human pose and shape accuracy, with the motion reconstruction and trajectory estimation accuracy in global coordinates.}
% MPJPE, PA-MPJPE, W-MPJPE, and WA-MPJPE are in $mm$, RTE is in \%.
%\mhc{make colors consistent. In lower tables, you have three colors for the three best methods. Here you sometimes have one color only or two per column.} 
% \RD{A bit counterintuitive that many numbers are higher for the regular terrain than irregular.}}
\label{table:recon}
\end{table*}

\begin{table*}[ht!]
\centering
\begin{tabular}{c|c|c|c|c|c|c}
\hline
& & & & \textbf{\% of frames with} & \textbf{Average penetration} & \textbf{Average distance} \\
\textbf{Scenes} & \textbf{Models} & \textbf{Jitter} $\downarrow$ & \textbf{FS} $\downarrow$ & \textbf{scene penetration} $\downarrow$ & \textbf{per frame} $\downarrow$ & \textbf{above the scene} $\downarrow$ \\
\hline
\multirow{4}{*}{\rotatebox[origin=c]{90}{\textbf{Non-flat}}}
& GLAMR \cite{Yuan2021GLAMRGO}      & 9.04 & 15.01 & \cellcolor{red!20}1.74 &\cellcolor{red!20}2.92 & 1370.82 \\
& WHAM \cite{Shin2023WHAMRW}        & \cellcolor{red!20}5.41& \cellcolor{orange!30}4.78& \cellcolor{orange!30}16.19  & \cellcolor{orange!30}40.23  &  1113.03 \\
& 4DHumans \cite{Goel2023HumansI4}  & \cellcolor{orange!30}7.29 & 13.99 & 84.56  & 192.73 & \cellcolor{red!20}285.54  \\
& \textbf{Ours}                     & 8.57 & \cellcolor{red!20}3.22 & 68.13 & 119.23  & \cellcolor{orange!30}377.37  \\
% & GT                                & 0.0 & 0.0 & 0.08 & 0.0016 & 162.11 \\
\hline
\multirow{4}{*}{\rotatebox[origin=c]{90}{\textbf{Flat}}} 
& GLAMR \cite{Yuan2021GLAMRGO}      & 6.88 & 10.00  & \cellcolor{red!20}0.0 & \cellcolor{red!20}0.0 &1694.46   \\
& WHAM \cite{Shin2023WHAMRW}        & \cellcolor{red!20}4.16 & \cellcolor{orange!30}3.96 & \cellcolor{orange!30}6.87  & \cellcolor{orange!30}8.38 & 2130.56  \\
& 4DHumans \cite{Goel2023HumansI4}  & \cellcolor{orange!30}5.19 & 9.33 & 97.92 & 180.96 & \cellcolor{orange!30}271.33  \\
& \textbf{Ours}                     & 6.89 & \cellcolor{red!20}1.80 & 34.24 & 86.56 & \cellcolor{red!20}196.17  \\
% & GT                                & 0.0 & 0.0 & 0.0 & 0.0 & 181.90 \\
\hline
\end{tabular}
\caption{Evaluating the physical plausibility of the methods with respect to the scene.}
% Jitter is in $mm/s$ and FS, average penetration per frame, average distance above the scene are in $mm$.
\label{table:physics}
\end{table*}

Table~\ref{table:physics} shows that our method minimizes foot sliding, improving stability. WHAM and GLAMR reduce scene penetration more than 4DHumans and our model, but their trajectories are often misaligned, positioning subjects unrealistically high, as seen in Figure~\ref{fig:pybullet_qualitative}.
Physics constraints improve plausibility but slightly reduce pose accuracy, a trade-off also observed in PhysCap~\cite{Shimada2020PhysCap}. Despite this, our model balances scene-aware motion and accurate pose estimation.
Metrics like W-MPJPE, WA-MPJPE, and RTE are higher on flat terrain due to SLAM errors from fewer visual features, increasing global trajectory errors.

% %-------------------------------------------------------------------------
\subsection{Ablation Studies}
To demonstrate the importance of the components in the enhanced physics module, we conduct ablation studies, selecting sequence $3$ on non-flat ground for these experiments. 
Results are in Tables~\ref{table:ablation_recon} and \ref{table:ablation_physics}.

\noindent \textbf{Only joint angle controller $\mathcal{E}_\theta$} yields low MPJPE and PA-MPJPE since joint angles match reference poses closely. 
However, lack of joint position supervision greatly increases W-MPJPE, WA-MPJPE, and causes significant scene penetration errors.

\noindent \textbf{Only joint position controller $\mathcal{E}_r$} results in low W-MPJPE and WA-MPJPE due to direct joint position supervision. 
However, MPJPE and PA-MPJPE increase from imprecise joint angle estimation, and physical plausibility errors, especially foot sliding, become significant.
These findings demonstrate the complementary roles of joint angle and position controllers.

\noindent \textbf{Flat scene without root supervision} tests the impact of using a flat floor instead of the height map and removing root supervision ("no drifting" term in Eq.~\eqref{eq:root_supervision}).
While MPJPE and PA-MPJPE remain similar, W-MPJPE and WA-MPJPE increase significantly, indicating inaccuracies in global positioning. 
Additionally, the incorrect global coordinates lead to an increased average distance of the subject above the scene.

\begin{table*}[ht!]
\centering
\begin{tabular}{c|c|c|c|c|c}
\hline
\textbf{Experiments} & \textbf{MPJPE} $\downarrow$ & \textbf{PA-MPJPE} $\downarrow$ & \textbf{W-MPJPE} $\downarrow$ & \textbf{WA-MPJPE} $\downarrow$ & \textbf{RTE}  $\downarrow$ \\
\hline
Only $\mathcal{E}_\theta $          & 192.32 & \cellcolor{red!20}61.56 & 1937.77 & 822.37 & 1.67 \\
Only $\mathcal{E}_r$                & 202.84 & 111.16 & \cellcolor{orange!30}508.59  & \cellcolor{red!20}354.58& \cellcolor{red!20}0.69 \\
w/o height map \& root supervision  & \cellcolor{orange!30}190.73 & \cellcolor{orange!30}61.72 & 1488.64 & 726.97 & 1.41  \\
\textbf{Ours}                       & \cellcolor{red!20}183.70 & 62.88 & \cellcolor{red!20}490.61 & \cellcolor{orange!30}359.32 & \cellcolor{orange!30}0.70  \\
\hline
\end{tabular}
\caption{Ablation study for different components of the physics module on evaluating the 3D human pose and shape accuracy, with the motion reconstruction and trajectory estimation accuracy in global coordinates.}
\label{table:ablation_recon}
\end{table*}

\begin{table*}[ht!]
\centering
\begin{tabular}{c|c|c|c|c|c}
\hline
& & & \textbf{\% of frames with} & \textbf{Average penetration} & \textbf{Average distance} \\
\textbf{Experiments} & \textbf{Jitter} $\downarrow$ & \textbf{FS} $\downarrow$ & \textbf{scene penetration} $\downarrow$ & \textbf{per frame}$\downarrow$ & \textbf{above the scene} $\downarrow$ \\
\hline
Only $\mathcal{E}_\theta$                       & 12.61 & 4.35 & 99.67 & 141.47 & 653.69  \\
Only $\mathcal{E}_r$                            & 10.19 & 7.75 & \cellcolor{red!20}51.25 & \cellcolor{red!20}59.27 & \cellcolor{orange!30}515.19    \\
w/o height map \& root supervision              & \cellcolor{red!20}8.24 & \cellcolor{orange!30}3.51 & 98.75 & 143.84 & 684.22 \\
\textbf{Ours}   & \cellcolor{orange!30}8.81 & \cellcolor{red!20}3.45 & \cellcolor{orange!30}87.12 & \cellcolor{orange!30}137.57 & \cellcolor{red!20}388.47 \\
\hline
\end{tabular}
\caption{Ablation study for different components of the physics module evaluating the physical plausibility.}
\label{table:ablation_physics}
\end{table*}

%
%%%%%%%%%%%%%%%%%%%%%%%%%%%%%%%%%%%
%
\section{Limitations and Future Work} Our benchmark, MoviCam, features only single-person sequences, with a pre-scanned, static scene where interactions are limited to foot-floor contact. 
Consequently, our method, PhysDynPose, focuses solely on foot-floor contact. 
%
% Since we rely on 4DHumans for pose estimation and DROID-SLAM for camera trajectory, PhysDynPose's performance is inherently limited by these tools and relies on ground truth camera trajectory for initialization.
Since PhysDynPose relies on 4DHumans for pose estimation and DROID-SLAM for camera trajectory, its performance is inherently limited by these tools and depends on ground truth camera initialization.
Future work could expand the benchmark to include more diverse human-scene interactions and include more diverse non-flat scenes while extending the physics module to monitor additional joints for contact. 
Additionally, instead of manually tuning PD controller gains, these parameters could be learned, as demonstrated in Neural PhysCap \cite{Shimada2021NeuralM3}.
%
%%%%%%%%%%%%%%%%%%%%%%%%%%%%%%%%%%%
%
\section{Conclusion}
In this study, we introduce a novel evaluation benchmark, MoviCam, specifically designed for human motion estimation in the world frame, providing a more detailed assessment than existing datasets. 
The evaluations conducted on our dataset demonstrate that the proposed benchmark is particularly effective in highlighting both the strengths and weaknesses of various methods. 
Alongside, we propose PhysDynPose, a scene-aware, physics-based method that estimates human motion in global coordinates by disentangling human motion from camera motion through a kinematics estimator and SLAM-derived camera trajectory. \
By optimizing global motion with physical constraints, including scene information, PhysDynPose achieves a balance between physical plausibility and motion accuracy, outperforming current approaches in reconstructing global trajectories.

{
    \small
    \bibliographystyle{ieeenat_fullname}
    \bibliography{main}
}


\end{document}